\documentclass[journal]{IEEEtran}
\usepackage{latexsym}

\usepackage{enumerate}
\usepackage{graphicx}
\usepackage{subfigure}
\usepackage{multirow}

\usepackage{framed,multirow}
\usepackage{times}
\usepackage{graphicx} % more modern
\usepackage{subfigure}
\usepackage{amssymb}
\usepackage{latexsym}
\newcommand{\tr}{\operatorname{tr}}
\usepackage{algorithm}
\usepackage{algorithmic}

\usepackage{hyperref}

\usepackage{url}
\usepackage{xcolor}
\usepackage{amsmath}
\usepackage{amsxtra}
\usepackage{braket}
\usepackage{array}
\usepackage{color}
\usepackage{url}
\usepackage{placeins}

\usepackage{times}
\usepackage{balance}
\usepackage{enumerate}
\usepackage{multirow}
\usepackage{pifont}
\usepackage{marvosym}
\usepackage{ifsym}
\usepackage{threeparttable}

\begin{document}
\title{HAQJSK: Hierarchical-Aligned Quantum Jensen-Shannon Kernels for Graph Classification}
\author{Lu~Bai,~\IEEEmembership{}
        Lixin~Cui,~\IEEEmembership{}
        Yue~Wang,~\IEEEmembership{}
        Ming~Li,~\IEEEmembership{}
        and~Edwin R.~Hancock,~\IEEEmembership{IEEE~Fellow}

\thanks{Lu Bai (E-mail: bailucs@cufe.edu.cn; bailu69@hotmail.com) is with School of Artificial Intelligence, Beijing Normal University, Beijing, China, and Central University of Finance and Economics, Beijing, China. Lixin Cui (${}^{*}$Corresponding Author: cuilixin@cufe.edu.cn), and Yue Wang are with Central University of Finance and Economics, Beijing, China. Ming Li is with Zhejiang Normal University, Zhejiang, China. Edwin R. Hancock is with Department
of Computer Science, University of York, York, UK}% <-this % stops a space
}
\markboth{IEEE Transactions on Neural Networks and Learning Systems}%
{Shell \MakeLowercase{\textit{et al.}}: Bare Demo of IEEEtran.cls for Journals}

\maketitle

\begin{abstract}
%\bol\begin{abstract}
In this work, we propose a family of novel quantum kernels, namely the Hierarchical Aligned Quantum Jensen-Shannon Kernels (HAQJSK), for un-attributed graphs. Different from most existing classical graph kernels, the proposed HAQJSK kernels can incorporate hierarchical aligned structure information between graphs and transform graphs of random sizes into fixed-sized aligned graph structures, i.e., the Hierarchical Transitive Aligned Adjacency Matrix of vertices and the Hierarchical Transitive Aligned Density Matrix of the Continuous-Time Quantum Walk (CTQW). For a pair of graphs to hand, the resulting HAQJSK kernels are defined by measuring the Quantum Jensen-Shannon Divergence (QJSD) between their transitive aligned graph structures. We show that the proposed HAQJSK kernels not only reflect richer intrinsic global graph characteristics in terms of the CTQW, but also address the drawback of neglecting structural correspondence information arising in most existing R-convolution kernels. Furthermore, unlike the previous Quantum Jensen-Shannon Kernels associated with the QJSD and the CTQW, the proposed HAQJSK kernels can simultaneously guarantee the properties of permutation invariant and positive definiteness, explaining the theoretical advantages of the HAQJSK kernels. Experiments indicate the effectiveness of the proposed kernels.
\end{abstract}


\maketitle
\IEEEpeerreviewmaketitle

\section{Introduction}\label{s1}

Graph-based structural data are powerful tools in various research domains that focus on modeling pairwise relationships between components, e.g., the analysis of social networks~\cite{DBLP:conf/pkdd/Bai0BH19}, molecule networks~\cite{DBLP:conf/nips/GasteigerBG21}, 3D shapes~\cite{DBLP:conf/cvpr/EscolanoHL11}, traffic networks~\cite{DBLP:journals/apin/ZengPHYH22}, etc. A central task of graph data analysis is how to learn significant numeric characteristics or features of the discrete graph structures for graph classification. To this end, one prevalent way is to adopt machine learning algorithms based on graph kernel methods, which can represent graph characteristics in a meaningful high dimensional Hilbert space and thus better reserve the structures~\cite{DBLP:journals/ftml/BorgwardtGLOR20}. This paper aims to propose a family of novel quantum information theoretic graph kernels associated with Continuous-Time Quantum Walks (CTQWs), for the purpose of graph classification. Our kernels are based on measuring the Quantum Jensen-Shannon Divergence (QJSD) between the hierarchical transitive aligned structures of pairwise graphs associated with the CTQW. The proposed kernels not only encapsulate transitive alignment information between graph structures, but also guarantee the positive definiteness.

\subsection{Literature Review}\label{s1.1}

Generally speaking, graph kernels are defined as the positive definite similarity measures between pairs of graphs~\cite{GarterCOLT2003,DBLP:conf/icml/KashimaTI03,DBLP:conf/cvpr/HarchaouiB07}. Perhaps the most successful and widely used way to define a graph kernel is the principle of R-convolution developed by Haussler~\cite{haussler99convolution}. It is a generic principle to define graph kernels between pairs of graphs by decomposing graph structures into substructures and measuring the isomorphism between the substructures, i.e., counting the pairs of isomorphic substructure between graphs. More specifically, one can employ any available graph decomposing algorithm to develop a graph kernel, e.g., the graph kernel based on comparing pairs of
decomposed a) paths~\cite{DBLP:journals/tnn/AzizWH13}, b) walks~\cite{DBLP:conf/nips/SugiyamaB15},  c) subtrees~\cite{DBLP:journals/jmlr/AzaisI20}, d) subgraphs~\cite{DBLP:conf/icml/KriegeM12}, etc.

Within the scenario of R-convolution, Kashima et al.~\cite{DBLP:conf/icml/KashimaTI03}
have defined a Random Walk-based Kernel by decomposing graph structures into random walks of different restricted lengths. The kernel is defined by calculating the number of the random walk pairs with different same lengths. Borgwardt et al.~\cite{DBLP:conf/icdm/BorgwardtK05} have developed a Shortest Path Kernel by counting the number of the shortest paths with different same lengths. Aziz et al.~\cite{DBLP:journals/tnn/AzizWH13} have developed a non-Backtrack Path Kernel through cycle-based structures that are abstracted from the Ihara zeta function~\cite{DBLP:journals/tnn/RenWH11}. The resulting kernel is computed by counting the numbers of cycle pairs with different same lengths. Costa and Grave~\cite{DBLP:conf/icml/CostaG10} have proposed a Neighborhood Subgraph
Pairwise Distance Kernel by decomposing graphs into layer-wise expansion neighborhood subgraphs rooted at a pair of vertices with specified distance and measuring the isomorphism between the subgraphs. Gaidon et al.~\cite{DBLP:conf/bmvc/GaidonHS11} have proposed a Subtree-based Kernel for video classification problems. They commence by representing complicated actions as spatio-temporal parts and transforming them into binary tree patterns. The proposed kernel is defined by comparing the pairs of isomorphic subtrees. Shervashidze et al.~\cite{shervashidze2010weisfeiler} have developed a Weisfeiler-Lehman Subtree Kernel by counting the pairs of isomorphic subtree patterns corresponded by the Weisfeiler-Lehman graph invariant. Other alternative R-convolution based graph kernels also include a) the Segmentation-based Graph Kernel~\cite{DBLP:conf/cvpr/HarchaouiB07}, b) the Pyramid Quantized Shortest Path Kernel~\cite{DBLP:journals/ijon/GkirtzouB16}, c) the Pyramid Quantized Weisfeiler-Lehman Subtree Kernel~\cite{DBLP:journals/ijon/GkirtzouB16}, d) the Wasserstein Weisfeiler-Lehman Kernel~\cite{DBLP:conf/nips/TogninalliGLRB19}, e) the Isolation Graph Kernel~\cite{DBLP:conf/aaai/XuTJ21}, f) the Graph Filtration Kernels~\cite{DBLP:conf/aaai/SchulzWW22}, g) the Attributed Subgraph Matching Kernel~\cite{DBLP:conf/icml/KriegeM12}, etc.

\begin{figure}
\centering
\subfigure{\includegraphics[width=1.0\linewidth]{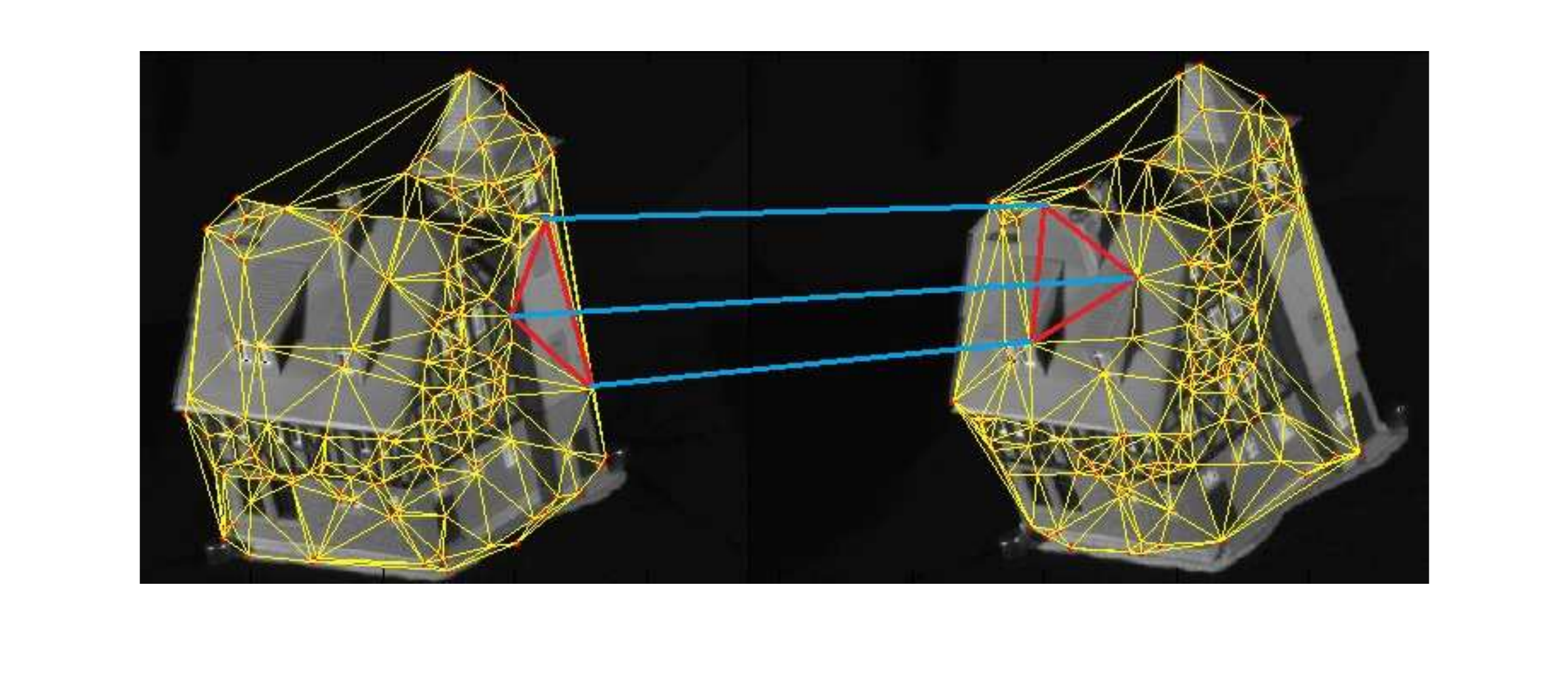}}
\vspace{-30pt}
\caption{\footnotesize{Graphs abstracted from digital images of different viewpoints.}} \label{cvpr}
\vspace{-20pt}
\end{figure}

One common shortcoming arising in most of the aforementioned R-convolution based kernels is that of ignoring the structural correspondence information between the graph structures. This problem is due to the fact that the R-convolution kernels only augment an unit kernel value if a pair of isomorphic substructures are detected, and the procedures of the isomorphism identification do not consider whether the isomorphic substructures are structurally aligned within the whole graph structures. For an instance of a computer vision problem in Fig.~\ref{cvpr}, there are a pair of graph structures extracted from two images, that contain the same house based on different viewpoints. The R-convolution kernels will directly augment an unit kernel value when they identify the isomorphic triangle-based substructures, no matter whether they are structurally aligned to each other through the vision background. Thus, the R-convolution kernels will not reflect the precise similarity measure between a pair of graphs, and may in turn influence the graph classification performance.

To address the above issues, Bai and Xu et al.~\cite{DBLP:conf/icml/Bai0ZH15,DBLP:conf/ijcai/BaiZW0H15,DBLP:journals/pr/XuBJTZL21} have developed a family of Depth-based Alignment Kernels that encapsulate the vertex correspondence information. For a pair of graphs, they first compute the Depth-based Complexity Trace rooted at each vertex as the vectorial vertex representations~\cite{DBLP:journals/pr/BaiH14}, the resulting alignment kernels are attained by counting the pairs of aligned vertices that are identified by evaluating the distance between the vectorial vertex representations in a Euclidean space. More specifically, these vertex alignment kernels are theoretically equivalent to aligned subgraph kernels that encapsulate structural correspondence information between substructures, addressing the shortcoming of neglecting structural correspondence information between graphs arising in existing R-convolution kernels.

Unfortunately, both the aforementioned R-convolution kernels and vertex alignment kernels fail to reflect characteristics of global graph structures. The reasons for this problem are twofold. First, the definition of the R-convolution kernels relies on the graph decomposition, that may cause the notorious computational inefficiency. Thus, the R-convolution kernels usually compromise to use substructures of small sizes, and it is hard to represent the global graph characteristics with such substructures. Second, the vertex alignment kernels only focus on identifying the correspondence information between local vertices, which only reflect local topological information. Moreover, on the other hand, the above alignment kernels are not positive definite. This is due to the fact that the aligned vertices identified by the alignment kernels are not transitive, i.e., if vertices $a$ and $b$ are both aligned to vertex $c$, we cannot guarantee that vertices $a$ and $b$ are also aligned to each other. The transitivity is a necessary condition to ensure the positive definiteness for an alignment or matching kernel~\cite{DBLP:conf/icml/FrohlichWSZ05}.

To overcome the drawback of neglecting global graph characteristics arising in the above R-convolution kernels and vertex alignment kernels, many graph kernels have been developed through the adjacency matrices that are natural representations of global graph structures. For example, Johansson et al.,~\cite{DBLP:conf/icml/JohanssonJDB14} have proposed a Global Graph Kernel based on the celebrated Lov\'asz numbers as well as their orthonormal representations computed through the adjacency matrices. Xu et al.,~\cite{DBLP:journals/pr/XuJBXL18} have developed a Hybrid Reproducing Kernel using the global entropy measures of the adjacency matrices.

Another alternative way of analyzing global graph structures is to employ the Continuous-time Quantum Walks (CTQW)~\cite{DBLP:journals/pr/EmmsWH09}. In quantum mechanics and quantum information theory~\cite{nielsen2010quantum}, the CTQW represents a quantum analogue of the classical Continuous-time Random Walk (CTRW), it is controlled by an unitary matrix and is not dominated by the low Laplacian spectrum frequencies. By contrast, the classical CTRW is controlled by a doubly stochastic matrix. Hence, the CTQW can better distinguish different global graphs. More specifically, there have been a family of Quantum Jensen-Shannon Kernels (QJSK) defined based on the CTQW. For example, Bai et al.~\cite{DBLP:journals/pr/Bai0TH15} have defined a quantum walk kernel by measuring the Quantum Jensen-Shannon Divergence (QJSD) between the density matrices of the CTQW evolved on graph structures. Rossi et al.~\cite{rossi2015measuring}, on the other hand, have proposed a quantum walk kernel by exploring the relationship between the interferences and symmetries of the CTQW evolved on graphs, in terms of the QJSD. Both the quantum walk kernels adopt the Laplacian matrix as the Hamiltonian operator, that can naturally reflect global graph characteristics. Unfortunately, computing the QJSD between a pair of graphs requires a composite graph structure that needs the vertex alignment information between the graphs, both the quantum kernels cannot guarantee the transitivity between the aligned vertices or even compute the composite structure by randomly arranging the vertex orders. Thus, these quantum kernels are not positive definite, and can not reflect the precise similarity between graphs. Broadly speaking, developing effective graph kernels is always a theoretical challenge problem.

\subsection{Contributions}

%The aim of this paper is to address the drawbacks of the aforementioned graph kernels by developing a family of new Hierarchical-Aligned Quantum Jensen-Shannon Kernels (HAQJSK) for Un-attributed Graphs. To this end, we develop our previous Quantum Jensen-Shannon Kernel (QJSK)~\cite{DBLP:journals/pr/Bai0TH15} one step further and theoretically generalize the QJSK kernel as a new Hierarchical Aligned QJSK kernel (i.e., the HAQJSK kernel), that can naturally guarantee the positive definiteness. One key innovation of the new HAQJSK kernels is to hierarchically employ the transitive vertex matching approach~\cite{DBLP:journals/tnnls/Cui21} and identify the transitive correspondence information between vertices of graphs.

The aim of this paper is to address the drawbacks of the aforementioned graph kernels by proposing a family of novel Hierarchical-Aligned Quantum Jensen-Shannon Kernels (HAQJSK), that not only encapsulate structural correspondence information of local vertices but also capture structural characteristics of global graph structures. To this end, we develop our previous Quantum Jensen-Shannon Kernel (QJSK)~\cite{DBLP:journals/pr/Bai0TH15} one step further and theoretically generalize the QJSK kernel as new hierarchical aligned QJSK kernels (i.e., the HAQJSK kernels). One of the key innovations for the proposed HAQJSK kernels is to identify the hierarchical transitive correspondence information between vertices of graphs. More specifically, the HAQSJK kernels can adopt the correspondence information to transform the original graphs of different sizes into fixed-sized hierarchical transitive aligned structures. The resulting HAQJSK kernels are defined by measuring the QJSD between the aligned structures of graphs associated with the CTQW. For the QJSD, since the required composite structure of a pair of graphs is computed based on their aligned graph structures and encapsulates the transitive correspondence information of the graphs, the HAQJSK kernels are theoretically equivalent to a transitive vertex alignment kernel and guarantee the positive definiteness. The contributions of this paper are threefold.

\textbf{First}, we introduce a framework to transform the arbitrary sized graphs into fixed-sized hierarchical transitive aligned structures, i.e., the hierarchical transitive aligned density matrix of the CTQW and the hierarchical transitive aligned adjacency matrix of the vertices. This is achieved by hierarchically adopting the transitive vertex matching approach~\cite{DBLP:journals/tnnls/Cui21} to identify the hierarchical transitive correspondence information between vertices of graphs, and the aligned structures are constructed through the associated hierarchical correspondence matrices. We show that the aligned structures not only remain the topological and quantum interference information residing on original graph structures, but also provide a natural way to define a transitive aligned QJSK kernel.

\textbf{Second}, with the hierarchical transitive aligned structures of graphs to hand, we develop a family of HAQJSK kernels by measuring the QJSD associated with the CTQW between the aligned structures. For a pair of graphs, since the required composite structure of the QJSD is computed by summing their aligned structure representations, the composite structure naturally takes into account the transitive vertex correspondences between the pair of graphs, reflecting the locational correspondences between the CTQW evolving on the graphs. Hence, the proposed HAQJSK kernels can be theoretically considered as the transitive aligned version of the previous QJSK kernels, that not only guarantee the positive definiteness but also reflect more precise kernel similarity measures than the existing QJSD based quantum kernels~\cite{DBLP:journals/pr/Bai0TH15,rossi2015measuring}. Furthermore, we show that the proposed HAQJSK kernels can simultaneously reflect the structural information residing on the local vertices as well as the global graph structures.

\textbf{Third}, we empirically demonstrate the effectiveness of the HAQJSK kernels associated with the C-SVMs for graph classification tasks. The HAQJSK kernels can outperform state-of-the-art graph kernels as well as graph neural network models on standard graph datasets.

% More specifically, similarly to the original ASGCN model, the proposed BASGCN model employs the transitive alignment procedure to transform arbitrary-sized graphs into fixed-sized aligned grid structures with consistent vertex orders, guaranteeing that the vertices on the same spatial position are also transitively aligned to each other in terms of the topological structures.

% \subsection{Paper Outline}
The remainder of this paper is organized as follows. Section \ref{s2} briefly reviews the background of the concepts of the CTQW and the previous QJSK kernel. Section \ref{s3} illustrates the concept for the new HAQJSK kernels. Section \ref{s4} exhibits the empirical results of the proposed HAQJSK kernels. Section \ref{s5} gives the conclusion of this work.

\section{Quantum Mechanical Backgrounds and Related Works}\label{s2}

% In this section, we briefly introduce the quantum mechanical backgrounds that will be used in this work. Furthermore, we review the definition of the previous QJSK kernel in the literatures. Theoretical analysis of the QJSK kernel is also provided in details.

In this section, we briefly review the quantum mechanical background that will be used in this work. Furthermore, we review the definition of the previous QJSK kernel in literatures.

% Theoretical analysis of the QJSK kernel is also provided in details.

\subsection{Continuous-time Quantum Walks}\label{s2.1}

In this subsection, we briefly review the concept of the CTQW. In quantum information theory~\cite{farhi1998quantum,kempe2003quantum,nielsen2010quantum}, the CTQW is the quantum analogue of the classical Continuous-Time Random Walk (CTRW) that simulates the transition process of a Markovian diffusion on the vertices of graphs through the adjacency matrices. Similar to the CTRW, the state space of the CTQW is the vertex set. More formally, given a sample graph $G(V,E)$, $V$ is the vertex set and $E$ is the edge set. With the Dirac notation, we define the basis state of the CTQW at vertex $u \in V$ as $\Ket{u}$, and $\Ket{.}$ is a $|V|$-dimensional orthonormal vector in a complex-valued Hilbert space $\mathcal{H}$. The state $\Ket{\psi(t)}$ of the CTQW at time $t$ is a linear combination of the orthonormal basis states $\Ket{u}$ over all vertices, i.e.,
\begin{equation}
\Ket{\psi(t)} = \sum_{u\in V} \alpha_u(t) \Ket{u}.
\end{equation}
Here $\alpha_u (t) \in \mathbb{C}$ is the complex amplitude of the basis state $\Ket{u}$ and follows the condition $\sum_{u \in V} \alpha_u (t) \alpha^{*}_u(t) = 1$ for all $u \in V$ and $t \in \mathbb{R}^{+}$, and $\alpha_u^* (t)$ is the complex conjugate of $\alpha_u (t)$. Different from the CTRW, we define the evolution of the CTQW using the Schr\"{o}dinger equation
\begin{equation}
\frac{ \partial \Ket{\psi_t} }{\partial t}  = -i\mathcal{H}\Ket{\psi_t},
\end{equation}
where $\mathcal{H}$ is the Hamiltonian of the system and specifies the total system energy. In this paper, we propose to adopt the Laplacian matrix $L$ as the Hamiltonian. More specifically, assume $A$ and $D$ are the vertex adjacency matrix and the vertex degree matrix of the graph $G$ respectively, $L$ is defined as $D-A$. Let $L=\Phi^T \Lambda \Phi$ be the spectral decomposition of $L$, where ${\Lambda}=diag({\lambda}_{1},{\lambda}_{2},...,{\lambda}_{{|{V}|}})$ is a diagonal matrix with the ordered eigenvalues as elements \(({\lambda}_{1}<{\lambda}_{2}<...<{\lambda}_{{|{V}|}})\) and
\({\Phi}=({\phi}_{1}|{\phi}_{2}|...|{\phi}_{{|{V}|}})\) is a matrix with the corresponding ordered orthonormal eigenvectors as columns. The state $\Ket{\psi(t)}$ of the CTQW at time $t$ can be computed as
\begin{equation}\label{stateFinal}
\Ket{\psi_t} = \Phi^T e^{-i \Lambda t} \Phi \Ket{\psi_0},
\end{equation}
where $\alpha_u(0)$ denotes an initial state and is computed by taking the square root of the vertex degree distribution of $G$~\cite{DBLP:journals/pr/Bai0TH15}.

Note that, the state $\Ket{\psi_t}$ at time $t$ is a pure state. However, we usually deal with the mixed state, that is a statistical ensemble of pure states $\Ket{\psi_t}$ each with a probability $p_t$ at time $t$. The density matrix (i.e., the density operator) of such a system is defined as $\rho = \sum_t p_t \Ket{\psi_t}\Bra{\psi_t}$, where $\Bra{\psi_t}$ is the conjugate transpose of $\Ket{\psi_t}$ and $\Ket{\psi_t}\Bra{\psi_t}$ is the outer product between them. Assume the CTQW evolves from time $t=0$ to time $t=T$. The mixed density matrix of $G$ can be defined as a time-averaged density matrix
\begin{equation}
\rho_G^T=\frac{1}{T} \int_{0}^T \! \Phi^T e^{-i\Lambda t} \Phi
\Ket{\psi_0}\Bra{\psi_0} \Phi^T e^{i\Lambda t} \Phi \, \mathrm{d}t. \label{MixDensityMatrix}
\end{equation}
Assume $\phi_{ra}$ and $\phi_{cb}$ are the $(ra)$-th and $(cb)$-th elements of the matrix of eigenvectors $\Phi$ of $L$. When $T$ approaches to $\infty$, we can employ the closed form solution to calculate the $(r,c)$-th element of $\rho_G^\infty$ as
\begin{equation}
\rho_G^\infty (r,c) = \sum_{\lambda \in \tilde{\Lambda}} \sum_{a \in B_{\lambda}} \sum_{b \in B_{\lambda}} \phi_{ra} \phi_{cb} \bar{\psi}_a \bar{\psi}_b,\label{FDensityOperator}
\end{equation}
where $\tilde{\Lambda}$ denotes the set of distinct eigenvalues of $L$, and $B_{\lambda}$ refers to the basis of the eigenspace associated with $\lambda$.

\vspace{5pt}

\noindent\textbf{Remarks:} The CTQW has some attractive properties that are not available for the classical CTRW. First, unlike the CTRW that has real-valued state vectors and is controlled by a stochastic matrix, the CTQW has complex-valued state vectors and its evolution is controlled by an unitary matrix. Thus, unlike the CTRW, the evolution of the CTQW is reversible and permits interference to take place, indicating that the CTQW can significantly reduce the drawback of tottering occurring in the CTRW. Moreover, the evolution of the CTQW is not governed by the low frequency components of the Laplacian spectrum. Hence, the CTQW can better discriminate various graph structures than the classical CTRW. Overall, the CTQW provides an elegant way to define novel algorithms for graph-based structure data analysis.

\subsection{The von Neumann Entropy}\label{s2.2}
In quantum information theory~\cite{nielsen2010quantum}, the von Neumann entropy $H_\mathrm{N}(\rho)$ of a quantum state (i.e., a density matrix) $\rho$ is formulated as
\begin{align}
H_\mathrm{N} = -\tr(\rho \log \rho)=-\sum_i \xi_i \ln \,\xi_i,
\end{align}
where $\xi_1,\ldots,\xi_n$ correspond to the eigenvalues of $\rho$. Generally, the von Neumann entropy of a pure state is zero, but the mixed state has a non-zero entropy value associated with its density matrix. In this work, we compute the von Neumann entropy of each graph $G$ associated with the density matrix $\rho_G^\infty$ of the CTQW defined in Eq.(\ref{FDensityOperator}). Specifically, the von Neumann entropy is defined as
\begin{equation}\label{vonEntropy}
H_\mathrm{N} (\rho_G) = -\tr(\rho_G \log{\rho_G}) = -\sum_j^{|V|} \lambda_j^G \log{\lambda_j^G},
\end{equation}
where $\lambda_j^G$ corresponds to the eigenvalues of $\rho_G^\infty$.

\subsection{The Quantum Jensen-Shannon Divergence}\label{s2.3}

In quantum information theory~\cite{lamberti2008metric}, the QJSD is defined as a quantum generalization of the classical Jensen-Shannon Divergence (JSD). Specifically, unlike the classical JSD defined to probability distributions, the QJSD is defined to quantum states. Assume a pair of density matrices $\rho$ and $\sigma$, the QJSD between them is formulated as
\begin{equation}\label{Eq:QJSD}
D_{\mathrm{QJS}}(\rho,\sigma) = H_\mathrm{N}\Big(\frac{\rho + \sigma}{2}\Big) -
\frac{1}{2} H_\mathrm{N}(\rho) - \frac{1}{2} H_\mathrm{N}(\sigma).
\end{equation}
Similar to the classical JSD, the QJSD is always well defined, symmetric, negative definite and bounded~\cite{lamberti2008metric}.

\subsection{The Quantum Jensen-Shannon Kernels for Graphs}\label{s2.4}
This subsection reviews the concepts for the QJSK kernel proposed in the previous work~\cite{DBLP:journals/pr/Bai0TH15}, that is also based on the QJSD associated with the CTQW. Assume we evolve the CTQW on a pair of graphs $G_p(V_p,E_p)$ and $G_q(V_q,E_q)$, and $\rho_{p}$ and $\rho_{q}$ are their associated mixed state density matrices defined by Eq.(\ref{FDensityOperator}). The QJSK kernel between $G_p$ and $G_q$ is defined as
\begin{equation}\label{UQJSGK}
k_{\mathrm{QJSU}}(G_p,G_q)=\mathrm{exp}\textbf{[}-\mu D_{\mathrm{QJS}}(\rho_p,\rho_q)\textbf{]},
\end{equation}
where
\begin{align}\label{Eq:QJSK}
D_{\mathrm{QJS}}(\rho_p,\rho_q) = H_\mathrm{N}\Big(\frac{\rho_p + \rho_q}{2}\Big)
- \frac{1}{2} H_\mathrm{N}(\rho_p) - \frac{1}{2} H_\mathrm{N}(\rho_q),
\end{align}
is the QJSD between the density matrices $\rho_{p}$ and $\rho_{q}$, $H_\textbf{N}(.)$ represents the von Neumann entropy given by Eq.(\ref{vonEntropy}), and $\mu$ is the decay factor and is usually set as $1$. Note that, since the sizes of the graphs $G_p$ and $G_q$ are usually different, straightforwardly computing their composite density matrix $\Big(\frac{\rho_p + \rho_q}{2}\Big)$ (i.e., a kind of composite structure of $G_p$ and $G_q$) tends to be elusive. One way to solve this problem is to expand the density matrix of the smaller graph with zero elements, so that the composite density matrix can be directly computed by summing the expanding density matrix of the smaller graph and the original density matrix of the larger graph. In fact, the QJSK kernel $k_{\mathrm{QJSU}}$ can be seen as an unaligned QJSK kernel without considering the row/column index orders (i.e., the vertex orders) between the density matrices, i.e., the kernel value of $k_{\mathrm{QJSU}}$ is not permutation invariant with the vertex orders. As a result, the QJSK kernel $k_{\mathrm{QJSU}}$ can not reflect precise similarity measures between graphs.

To address the permutation invariant problem arising in the QJSK kernel $k_{\mathrm{QJSU}}$, Bai et al.~\cite{DBLP:journals/pr/Bai0TH15} propose to compute the composite density matrix after the original density matrices are optimally aligned, i.e., compute lower bound of the QJSD over the state permutation set $\Sigma$. This in turn results an aligned QJSK kernel between $G_p$ and $G_q$ as
\begin{align}\label{AQJSGK}
k_{\mathrm{QJSA}}(G_p,G_q)&=\max_{Q\in \Sigma} \exp\textbf{[}-\mu D_{\mathrm{QJS}}(\rho_p,Q\rho_q Q^T)\textbf{]} \nonumber \\
&=\exp\textbf{[}-\mu \min_{Q\in \Sigma} D_{\mathrm{QJS}}(\rho_p,Q\rho_q Q^T)\textbf{]},
\end{align}
where $Q\in \{0,1\}^{|V_p|\times |V_q|}$ ($|V_p| \geq |V_q|$) is the vertex correspondence matrix between $G_p$ and $G_q$. Here, the correspondence matrix $Q$ is computed based on the Umeyama Spectral Matching Method~\cite{DBLP:journals/pami/Umeyama88}, that directly utilizes the eigendecomposition of the density matrices. More specifically, it is clear that vertices $u\in V_p$ and $v\in V_q$ are aligned to each other, if the $(u,v)$-th element of $Q$ is $1$.

\begin{figure*}[t]
\centering
\subfigure{\includegraphics[width=1.0\linewidth]{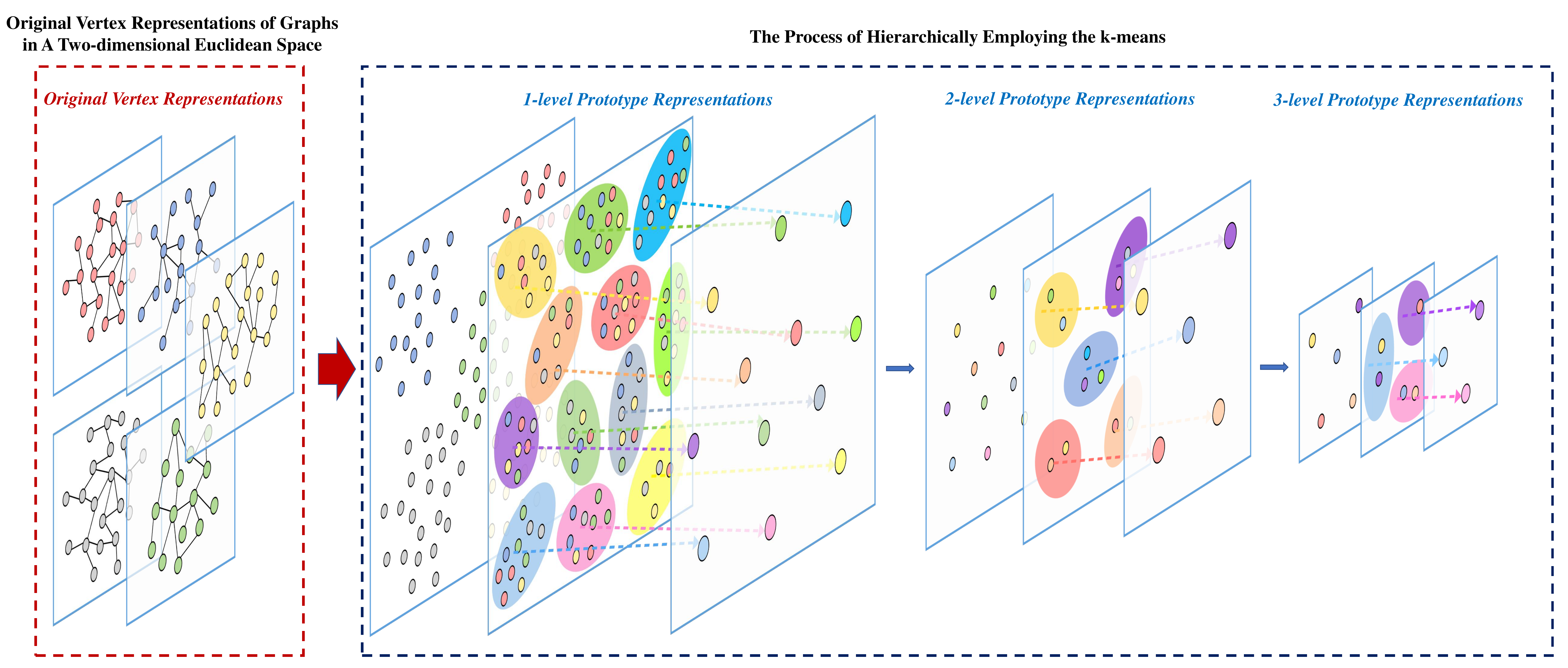}}
\vspace{-15pt}
\caption{\footnotesize{The process of computing the family of $2$-level hierarchical prototype representations. Assume there are five graphs, we first use their $k$-dimensional (i.e., $k=2$) vectorial representations as the $0$-level prototype representations. By performing $\kappa$-means, we can hierarchically locate a number of mean points as the $h$-level prototype representations over the $h$-level prototype representations. Here, $h$ varies from 1 to 3.}} \label{HPR_framework}
\vspace{-15pt}
\end{figure*}

\vspace{5pt}

\noindent\textbf{Remarks:} Both the QJSK kernels $k_{\mathrm{QJSU}}$ and $k_{\mathrm{QJSA}}$ can pad the density matrix of the smaller graph to the same size of the larger graph. This process is equivalent to extending the adjacency matrix of the smaller graph to the same size of the larger graph. Since the density matrix of the CTQW can encapsulate more complicated structure information than the original adjacency matrix, and the required von Neumann entropy of the density matrix can be seen as a global characteristics of a graph structure. As a result, comparing to the state-of-the-art graph kernels reviewed in Section~\ref{s1.1}, the QJSK kernels not only reflect more global structural information than most existing R-convolution kernels that compromise to use substructures of limited sizes, but also better discriminate different graph structures than most existing graph kernels that focus on capturing global characteristics through the adjacency matrix. Unfortunately, both the QJSK kernels $k_{\mathrm{QJSU}}$ and $k_{\mathrm{QJSA}}$ still suffer from a number of theoretical drawbacks. First, the unaligned QJSK kernel $k_{\mathrm{QJSU}}$ does not encapsulate any correspondence information between vertices into the computation of the required QJSD. Thus, $k_{\mathrm{QJSU}}$ cannot guarantee the positive definiteness and is not permutation invariant to the vertex orders. Second, although the aligned QJSK kernel $k_{\mathrm{QJSA}}$ employs the Umeyama matching method to identify the vertex correspondence information between pairs of graphs, and thus overcomes the permutation invariant problem. The aligned vertices are not guaranteed to be transitive. As a result, $k_{\mathrm{QJSA}}$ is also not a positive definite kernel. This indicates that both the unaligned and aligned QJSK kernels $k_{\mathrm{QJSU}}$ and $k_{\mathrm{QJSA}}$ cannot reflect a precise similarity measure between graphs. As an evidence, the experimental evaluations in ~\cite{DBLP:journals/pr/Bai0TH15} demonstrate that the classification performance of the C-SVMs associated with the aligned QJSK kernel $k_{\mathrm{QJSA}}$ is not significantly better than that associated with the unaligned QJSK kernel $k_{\mathrm{QJSU}}$. In this paper, we will propose a new variant of the HAQJSK kernel to overcome the above problems.

% thus computing  a lower bound of the divergence over the set $\Sigma$ of state permutations.
% \textbf{Remarks:}

\section{The Hierarchical Aligned Quantum Jesen-Shannon Graph Kernels}\label{s3}

In this section, we define a family of Hierarchical Aligned Quantum Jensen-Shannon Kernels (HAQJSK) for un-attributed graphs. We commence by transforming random sized graphs into fixed-sized hierarchical transitive aligned structures, through a hierarchical vertex matching method. Moreover, for a pair of graphs, we define the HAQJSK kernels by measuring the QJSD between their aligned structures. Finally, we indicate the theoretical advantages of the proposed HAQJSK kernels, explaining the effectiveness.

\subsection{Hierarchical Transitive Aligned Graph Structures}\label{s3.1}

In this subsection, we develop a framework to convert a set of graphs $\mathbf{G}$ with arbitrary sizes into fixed-sized hierarchical transitive aligned structures. To this end, we commence by identifying the hierarchical transitive correspondence information between graphs in $\mathbf{G}$, by hierarchically using the transitive vertex matching approach proposed in~\cite{DBLP:journals/tnnls/Cui21}. Assume the $k$-dimensional vectorial representations of the $N$ vertices $\mathbf{V}$ over all graphs in $\mathbf{G}$ are denoted as
\begin{equation}
\mathbf{{R}}^k(\mathbf{V}) =\{\mathrm{R}^k(v_1),\mathrm{R}^k(v_2),\ldots,\mathrm{R}^k(v_i),\ldots,\mathrm{R}^k(v_N)\},
\end{equation}
where $|\mathbf{V}|=N$ and $v_i\in \mathbf{V}$. To identify the transitive vertex correspondence information, Cui et al.~\cite{DBLP:journals/tnnls/Cui21} have aligned the vertices of each individual graph in $\mathbf{G}$ to a common set of $k$-dimensional prototype representations  $\mathbf{{P}}^k(\mathbf{G})$ over all graphs in $\mathbf{G}$. This is done by employing the classical $\kappa$-means clustering method on $\mathbf{{R}}^k(\mathbf{V})$ to locate a set of $k$-dimensional mean vectors, through minimizing the following objective function
\begin{equation}
\arg\min_{\Omega}  \sum_{j=1}^{N} \sum_{\mathrm{R}^k(v_i) \in c_j} \|\mathrm{R}^k(v_i)- \mu_j^k \|^2_2,\label{kmeans}
\end{equation}
where $\Omega=(c_1,c_2,\ldots,c_j,\ldots,c_{N})$ corresponds to $N$ clusters over $\mathbf{{R}}^k(\mathbf{V})$, and $\mu_j^k$ is the mean of all $\mathrm{R}^k(v_i)$ belonging to the $j$-th cluster $c_j$. As a result, the set of the $k$-dimensional prototype representations $\mathbf{{P}}^k(\mathbf{G})$ are defined as
\begin{equation}
\mathbf{{P}}^k(\mathbf{G}) =\{\mu_1^k,\mu_2^k,\ldots,\mu_j^k,\ldots,\mu_{N}^k\}.\label{PR}
\end{equation}
For a graph $G_p(V_p,E_p)\in \mathbf{G}$ , we align its $k$-dimensional vectorial vertex representations $\mathbf{{R}}^k(V_p)$ to $\mathbf{{P}}^k(\mathbf{G})$ and compute the correspondence matrix $C_p^k\in{0,1}^{|V_p| \times |\mathbf{{P}}^k(\mathbf{G})|}$ as
\begin{equation}
C^{k}_{p}(i,j)=\left\{
\begin{array}{cl}
1   & \mathrm{if} \ \mathrm{R}^k(v_i)\in \mathbf{{R}}^k(V_p)  \ \mathrm{of} \ v_i\in V_p \ \mathrm{belongs \ to}  \\
    & \mathrm{cluster} \ c_j\in \Omega, \ \mathrm{i.e., \ vertex} \ v_i\in V_p \ \mathrm{is}\\
    & \mathrm{aligned \ to} \ \mu_j^k\in \mathbf{{P}}^k(\mathbf{G}) \\
0   & \mathrm{otherwise}.
\end{array} \right.
\label{CoMatrix1}
\end{equation}
If $C^{k}_{p}(i,j)=1$, there exists an one-to-one correspondence information between the $i$-th vertex $v_i\in V_p$ and the $j$-th prototype representation $\mu_j^k\in \mathbf{{P}}^k(\mathbf{G})$. Because $\mu_j^k$ is the nearest prototype representation to $v_i$ based on Eq.(\ref{kmeans}). More specifically, assume a pair of graphs $G_p$ and $G_q$ from $\mathbf{G}$, if their vertices $v_i\in V_p$ and $v_j\in V_q$ are aligned to the same prototype representation $\mu_j^k\in \mathbf{{P}}^k(\mathbf{G})$, we think that the vertices $v_p$ and $v_q$ are transitively aligned to each other.

To further reflect the multi-scale structural information, we propose to hierarchically employ the above matching method and identify the hierarchical correspondence information. Assume $\mathbf{{R}}^k(\mathbf{V})$ is the set of $0$-level prototype representations $\mathbf{{P}}^{0,k}$. By hierarchically performing the $\kappa$-means on the $h-1$-level prototype representations $\mathbf{{P}}^{h-1,k}$, we compute a family of different $h$-level prototype representations over the vertices of all graphs in $\mathbf{G}$ as
\begin{equation}
\mathbf{HP}^{H,k}(\mathbf{G})=\{\mathbf{{P}}^{1,k}(\mathbf{G}),\ldots,\mathbf{{P}}^{h,k}(\mathbf{G}),\ldots,\mathbf{{P}}^{H,k}(\mathbf{G})\},\label{HPR}
\end{equation}
where $1 \leq h \leq H$, and $\mathbf{{P}}^{1,k}(\mathbf{G})$ is essentially $\mathbf{{P}}^{k}(\mathbf{G})$ defined in Eq.(\ref{PR}). Fig.\ref{HPR_framework} exhibits a detailed example of computing a family of different $h$-level prototype representations $\mathbf{HP}^{H,k}(\mathbf{G})$ defined in Eq.(\ref{HPR}). Moreover, we align each graph $G_p\in \mathbf{G}$ to each $h$-level prototype representations $\mathbf{{P}}^{h,k}\in \mathbf{HP}^{H,k}(\mathbf{G})$ and compute a family $h$-level hierarchical correspondence matrices as
\begin{equation}
{\mathbf{C}}^{H,k}(G_p)=\{ C^{1,k}_{p}  , C^{2,k}_{p} ,\ldots,C^{h,k}_{p},\ldots,C^{H,k}_{p}    \},\label{HC}
\end{equation}
where each correspondence matrix $C^{h,k}_{p} \in \{0,1\}^{|V_p|\times |\mathbf{{P}}^{h,k}|}$, and $C^{h,k}_{p}$ is essentially $C_p^k$ defined in Eq.(\ref{CoMatrix1}).

With the family of $h$-level hierarchical correspondence matrices $C^{h,k}_{p}$ to hand, we transform each graph $G_p\in \mathbf{G}$ into the fixed-sized transitive aligned structures through the correspondence information. Specifically, for each graph $G_p$, assume $A_p$ is its vertex adjacency matrix, and $\rho_p$ is its density matrix associated with the CTQW defined in Eq.(\ref{FDensityOperator}). We transform $G_p$ into a family of fixed-sized $h$-level transitive aligned adjacency matrices as
\begin{equation}
\mathcal{A}=\{ A_p^{1,k} , A_p^{2,k} ,\ldots,A_p^{h,k},\ldots,A_p^{H,k}\},\label{kHAA}
\end{equation}
where each $h$-level transitive aligned adjacency matrix $A_p^{h,k}\in \mathbb{R}^{|\mathbf{P}^{h,k}|\times |\mathbf{P}^{h,k}|}$ is
\begin{equation}
A_p^{h,k}={C^{1,k}_{p}}^{\mathrm{T}} A_p {C^{h,k}_{p}}.\label{AAM}
\end{equation}
Similarly, we transform $G_p$ into a family of fixed-sized $h$-level transitive aligned density matrices as
\begin{equation}
\mathcal{P}_p=\{ \rho_p^{1,k} , \rho_p^{2,k} ,\ldots,\rho_p^{h,k},\ldots,\rho_p^{H,k}\},\label{kHAA}
\end{equation}
where each $h$-level transitive aligned density matrix $\rho_p^{h,k}\in \mathbb{R}^{|\mathbf{P}^{h,k}|\times |\mathbf{P}^{h,k}|}$ is defined as
\begin{equation}
\rho_p^{h,k}={C^{1,k}_{p}}^{\mathrm{T}} \rho_p {C^{h,k}_{p}}.\label{AAD}
\end{equation}

Finally, note that, we use the $k$-dimensional depth-based (DB) representations as the vectorial vertex representations $\mathbf{{R}}^k(\mathbf{V})$, following the statement in~\cite{DBLP:journals/tnnls/Cui21}. It has been shown that the DB representation can preserve meaningful entropy flows rooted from each local vertex to the global graph structure through a family of $k$-layer expansion subgraphs rooted at the local vertex~\cite{DBLP:journals/pr/BaiH14,DBLP:journals/pr/BaiEH16}. To encapsulate more nested structural information, we can vary $k$ from $1$ ro $K$ (i.e., the largest layer of the expansion subgraphs, that corresponds to the greatest shortest path length over all graphs in $\mathbf{G}$), and thus compute the family of \textbf{Hierarchical Transitive Aligned Adjacency Matrices} as
\begin{equation}
\mathcal{\bar{A}}_p=\{ \bar{A}_p^{1} , \bar{A}_p^{2} ,\ldots, \bar{A}_p^{h},\ldots,\bar{A}_p^{H}\},\label{HAA}
\end{equation}
where
\begin{equation}
\bar{A}_p^{h}=\sum_{k=1}^{K} \frac{A_p^{h,k}}{K}.\label{HAAs}
\end{equation}
similarly, we also compute the family of \textbf{Hierarchical Transitive Aligned Density Matrices} as
\begin{equation}
\mathcal{\bar{P}}_p=\{ \bar{\rho}_p^{1} , \bar{\rho}_p^{2} ,\ldots, \bar{\rho}_p^{h},\ldots,\bar{\rho}_p^{H}\},\label{HAD}
\end{equation}
where
\begin{equation}
\bar{\rho}_p^{h}=\sum_{k=1}^{K} \frac{\rho_p^{h,k}}{K}.\label{HADs}
\end{equation}

\noindent\textbf{Remarks:} Clearly, both the Hierarchical Transitive Aligned Adjacency Matrix $\bar{A}_p^{h}\in \mathcal{\bar{A}}$ and the Hierarchical Transitive Aligned Density Matrix $\bar{\rho}_p^{h}\in \mathcal{\bar{P}}$ encapsulate the transitive aligned vertex correspondence information between all graphs in $\mathbf{G}$, i.e., for any pair of graphs in $\mathbf{G}$, their aligned adjacency matrices or aligned density matrices are transitively aligned to each other. This is because $\bar{A}_p^{h}\in \mathcal{\bar{A}}$ and $\bar{\rho}_p^{h}\in \mathcal{\bar{P}}$ are both computed by transforming the adjacency matrix and the density matrix of original graphs through the hierarchical correspondence matrices defined by Eq.(\ref{HC}), and their rows and columns correspond to the same hierarchical prototype representations defined by Eq.(\ref{HPR}). As a result, the aligned structures $\mathcal{\bar{A}}$ and $\mathcal{\bar{P}}$ provide a natural way to define new transitive aligned quantum Jensen-Shannon graphs kernels by measuring the QJSD between the aligned structures.

\subsection{The Proposed HAQJSK Kernels}
In this subsection, we define a family of HAQJSK kernels between graphs in $\mathbf{G}$ through the hierarchical aligned graph structures defined in Section~\ref{s3.1}.

% Specifically, assume a pair graphs $G_p\in \mathbf{G}$ and $G_q\in \mathbf{G}$, $$\mathcal{\bar{A}}_p=\{ \bar{A}_p^{1} , \bar{A}_p^{2} ,\ldots, \bar{A}_p^{h},\ldots,\bar{A}_p^{H}\}$$ and $$\mathcal{\bar{A}}_q=\{ \bar{A}_q^{1} , \bar{A}_q^{2} ,\ldots, \bar{A}_q^{h},\ldots,\bar{A}_q^{H}\}$$ are their associated sets of fixed-sized Hierarchical Aligned Adjacency Matrices defined by Eq.(\ref{HAA}), and $$\mathcal{\bar{P}}_p=\{ \bar{\rho}_p^{1} , \bar{\rho}_p^{2} ,\ldots, \bar{\rho}_p^{h},\ldots,\bar{\rho}_p^{H}\}$$ and $$\mathcal{\bar{P}}_q=\{ \bar{\rho}_q^{1} , \bar{\rho}_q^{2} ,\ldots, \bar{\rho}_q^{h},\ldots,\bar{\rho}_q^{H}\}$$ are their associated sets of Hierarchical Aligned Density Matrices defined by Eq.(\ref{HAD}). The HAQJSK kernels are defined as follows.

\vspace{5pt}

\noindent \textbf{Definition 3.1 (The HAQJSK Kernel based on Hierarchical Transitive Aligned Adjacency Matrices)}: For the pair of graphs $G_p\in \mathbf{G}$ and $G_q\in \mathbf{G}$ defined previously, let $$\mathcal{\bar{A}}_p=\{ \bar{A}_p^{1} , \bar{A}_p^{2} ,\ldots, \bar{A}_p^{h},\ldots,\bar{A}_p^{H}\}$$ and $$\mathcal{\bar{A}}_q=\{ \bar{A}_q^{1} , \bar{A}_q^{2} ,\ldots,\bar{A}_q^{h},\ldots,\bar{A}_q^{H}\}$$ denote their associated families of fixed-sized Hierarchical Transitive Aligned Adjacency Matrices defined by Eq.(\ref{HAA}). We commence by employing Eq.(\ref{FDensityOperator}) to compute the density matrices of the CTQW evolving on the aligned adjacency matrices in $\mathcal{\bar{A}}_p$ and $\mathcal{\bar{A}}_q$ as
$$\mathcal{\bar{Q}}_{p}=\{ \bar{\theta}_{p}^{1} , \bar{\theta}_{p}^{2} ,\ldots, \bar{\theta}_{p}^{h},\ldots,\bar{\theta}_{p}^{H}\}$$ and $$\mathcal{\bar{Q}}_{q}=\{ \bar{\theta}_{q}^{1} , \bar{\theta}_{q}^{2} ,\ldots, \bar{\theta}_{q}^{h},\ldots,\bar{\theta}_{q}^{H}\}.$$
The HAQJSK kernel $K_{\mathrm{HAQJS}}^{\mathrm{A}}$ between $G_p$ and $G_q$ associated with $\mathcal{\bar{A}}_p$ and $\mathcal{\bar{A}}_q$ is defined as
\begin{align}
K_{\mathrm{HAQJS}}^{\mathrm{A}}(G_p,G_q)&=K_{\mathrm{HAQJS}}^{\mathrm{A}}(\mathcal{\bar{Q}}_{p},\mathcal{\bar{Q}}_{q})
\nonumber \\
&=\sum_{h=1}^{H}\exp\textbf{[}-D_{\mathrm{QJS}}(\bar{\theta}_{p}^{h},\bar{\theta}_{q}^{h})\textbf{]},
\end{align}
where
\begin{align}
D_{\mathrm{QJS}}(\bar{\theta}_{p}^{h},\bar{\theta}_{q}^{h})&= H_\mathrm{N}(\frac{\bar{\theta}_{p}^{h}+\bar{\theta}_{q}^{h}}{2})-\frac{1}{2}H_\mathrm{N}(\bar{\theta}_{p}^{h})-\frac{1}{2}H_\mathrm{N}(\bar{\theta}_{q}^{h}).\label{QJSDA}
\end{align}
Let $\delta(A): A\rightarrow \rho$ represents the mapping function (i.e., Eq.(\ref{FDensityOperator})) that computes the density matrix $\rho$ of the CTQW evolving on the adjacency matrix $A$ of a graph $G$. Based on Eq.(\ref{AAM}) and Eq.(\ref{HAAs}), Eq.(\ref{QJSDA}) is further written as
\begin{align}
&D_{\mathrm{QJS}}(\bar{\theta}_{p}^{h},\bar{\theta}_{q}^{h})= D_{\mathrm{QJS}}(\delta(\bar{A}_p^h),\delta(\bar{A}_p^h))\nonumber \\
&= D_{\mathrm{QJS}}\textbf{[}\delta(   \frac{ \Sigma_{k=1}^K {C^{h,k}_{p}}^{\mathrm{T}} A_p {C^{h,k}_{p}} }{K}  ),\delta( \frac{ \Sigma_{k=1}^K {C^{h,k}_{q}}^{\mathrm{T}} A_q {C^{h,k}_{q}} }{K}  ) \textbf{]}\nonumber \\
&=H_\mathrm{N}\textbf{[}\frac{\delta(   \frac{ \Sigma_{k=1}^K {C^{h,k}_{p}}^{\mathrm{T}} A_p {C^{h,k}_{p}} }{K}  ) + \delta( \frac{ \Sigma_{k=1}^K {C^{h,k}_{q}}^{\mathrm{T}} A_q {C^{h,k}_{q}} }{K}}{2})\textbf{]}\nonumber \\
&\ \ \ -\frac{1}{2}H_\mathrm{N}\textbf{[}\delta(   \frac{ \Sigma_{k=1}^K {C^{h,k}_{p}}^{\mathrm{T}} A_p {C^{h,k}_{p}} }{K}  )\textbf{]}\nonumber \\
&\ \ \ -\frac{1}{2}H_\mathrm{N}\textbf{[}\delta(   \frac{ \Sigma_{k=1}^K {C^{h,k}_{q}}^{\mathrm{T}} A_q {C^{h,k}_{q}} }{K}  )\textbf{]},\label{QJSDA_R}
\end{align}
where $A_p$ and $A_q$ are the original adjacency matrices of $G_a$ and $G_p$, and $C^{h,k}_{p}$ and $C^{h,k}_{q}$ are the $h$-level correspondence matrices of $G_a$ and $G_p$. \hfill$\blacksquare$
% \label{QJSDA}

% \label{MixDensityMatrix}

% \label{FDensityOperator}

\vspace{5pt}
% Transitive Aligned
\noindent \textbf{Definition 3.2 (The HAQJSK Kernel based on Hierarchical Transitive Aligned Density Matrices)}: For the pair of graphs $G_p\in \mathbf{G}$ and $G_q\in \mathbf{G}$ defined previously, let $$\mathcal{\bar{P}}_p=\{ \bar{\rho}_p^{1} , \bar{\rho}_p^{2} ,\ldots, \bar{\rho}_p^{h},\ldots,\bar{\rho}_p^{H}\}$$ and $$\mathcal{\bar{P}}_q=\{ \bar{\rho}_q^{1} , \bar{\rho}_q^{2} ,\ldots,\bar{\rho}_q^{h},\ldots,\bar{\rho}_q^{H}\}$$ denote their associated families of fixed-sized Hierarchical Transitive Aligned Density Matrices defined by Eq.(\ref{HAD}). The HAQJSK kernel $K_{\mathrm{HAQJS}}^{\mathrm{D}}$ between $G_p$ and $G_q$ associated with $\mathcal{\bar{A}}_p$ and $\mathcal{\bar{A}}_q$ is defined as
\begin{align}
K_{\mathrm{HAQJS}}^{\mathrm{D}}(G_p,G_q)&=K_{\mathrm{HAQJS}}^{\mathrm{D}}(\mathcal{\bar{P}}_p,\mathcal{\bar{P}}_q)\nonumber \\
&=\sum_{h=1}^{H}\exp(-D_{\mathrm{QJS}}(\bar{\rho}_{p}^{h},\bar{\rho}_{q}^{h})),
\end{align}
where
\begin{align}
D_{\mathrm{QJS}}(\bar{\rho}_{p}^{h},\bar{\rho}_{q}^{h})&= H_\mathrm{N}(\frac{\bar{\rho}_{p}^{h}+\bar{\rho}_{q}^{h}}{2})-\frac{1}{2}H_\mathrm{N}(\bar{\rho}_{p}^{h})-\frac{1}{2}H_\mathrm{N}(\bar{\rho}_{q}^{h}).\label{QJSDD}
\end{align}
Based on Eq.(\ref{AAM}) and Eq.(\ref{HADs}), Eq.(\ref{QJSDD}) is further written as
\begin{align}
&D_{\mathrm{QJS}}(\bar{\rho}_{p}^{h},\bar{\rho}_{q}^{h})=D_{\mathrm{QJS}}(\sum_{k=1}^{K} \frac{\rho_p^{h,k}}{K},\sum_{k=1}^{K} \frac{\rho_q^{h,k}}{K})\nonumber \\
&=H_\mathrm{N}(\frac{\sum_{k=1}^{K} \frac{{C^{1,k}_{p}}^{\mathrm{T}} \rho_p {C^{h,k}_{p}}}{K} + \sum_{k=1}^{K} \frac{{C^{1,k}_{q}}^{\mathrm{T}} \rho_q {C^{h,k}_{q}}}{K} }{2})\nonumber \\
&\ \ \ -\frac{1}{2}H_\mathrm{N}(\sum_{k=1}^{K} \frac{{C^{1,k}_{p}}^{\mathrm{T}} \rho_p {C^{h,k}_{p}}}{K})-\frac{1}{2}H_\mathrm{N}(\sum_{k=1}^{K} \frac{{C^{1,k}_{q}}^{\mathrm{T}} \rho_q {C^{h,k}_{q}}}{K}),\label{QJSDD_R}
\end{align}
where $\rho_p$ and $\rho_q$ are the original density matrices of the CTQW evolving on $G_p$ and $G_q$, and $C^{h,k}_{p}$ and $C^{h,k}_{q}$ are the $h$-level correspondence matrices of $G_a$ and $G_p$.\hfill$\blacksquare$

\vspace{5pt}

\noindent\textbf{Lemma.} \emph{The HAQJSK kernels $K_{\mathrm{HAQJS}}^{\mathrm{A}}$ and $K_{\mathrm{HAQJS}}^{\mathrm{D}}$ is positive definite (\textbf{pd}).}

\vspace{5pt}

\noindent\textbf{Proof.} The reasons of the positive definiteness for the proposed HAQJSK kernels $K_{\mathrm{HAQJS}}^{\mathrm{A}}$ and $K_{\mathrm{HAQJS}}^{\mathrm{D}}$ are twofold. First, the associated QJSD measure for the HAQJSK kernels is a symmetric dissimilarity measure~\cite{majtey2005jensen}. Thus, the HAQJSK kernels defined as the negative exponential of the QJSD measure will be positive definite, if they are transitive aligned kernels~\cite{DBLP:conf/icml/FrohlichWSZ05,DBLP:journals/pr/BaiH14B}, following the theoretical statement in~\cite{DBLP:journals/pr/NeuhausB06,DBLP:journals/pr/Bai0TH15}. Second, Eq.(\ref{QJSDA_R}) and Eq.(\ref{QJSDD_R}) indicate that the $h$-level correspondence matrices $C^{h,k}_p$ and $C^{h,k}_q$ can transform the original adjacency matrices or density matrices of any pair of graphs $G_p$ and $G_q$ into fixed-sized aligned structures. Since $C^{h,k}_p$ and $C^{h,k}_q$ are computed by aligning each graph to a common $h$-level prototype representations, the correspondence information identified by $C^{h,k}_p$ and $C^{h,k}_q$ are transitive. Thus, the transformed aligned structures of different pairs of graphs are also transitively aligned to each other. Moreover, for Eq.(\ref{QJSDA_R}) and Eq.(\ref{QJSDD_R}), since the required composite structures $$\textbf{[}\frac{\delta(   \frac{ \Sigma_{k=1}^K {C^{h,k}_{p}}^{\mathrm{T}} A_p {C^{h,k}_{p}} }{K}  )+\delta( \frac{ \Sigma_{k=1}^K {C^{h,k}_{q}}^{\mathrm{T}} A_q {C^{h,k}_{q}} }{K}}{2})\textbf{]}$$ and $$(\frac{\sum_{k=1}^{K} \frac{{C^{1,k}_{p}}^{\mathrm{T}} \rho_p {C^{h,k}_{p}}}{K} + \sum_{k=1}^{K} \frac{{C^{1,k}_{q}}^{\mathrm{T}} \rho_q {C^{h,k}_{q}}}{K} }{2})$$ are computed by summing the aligned structures, these composite structures are also transitively aligned to the aligned structures of different graphs. Thus, the proposed HAQJSK kernels can be seen as transitive aligned kernels, that compute the similarity between transitive aligned structures. In summary, the proposed HAQJSK kernels are positive definite. \hfill$\Box$

% is a symmetric dissimilarity measure [48]. Since the kernel is computed as the negative exponential of the divergence measure, it follows that it is positive definite

\vspace{5pt}

\noindent\textbf{Remarks:} Since the proposed HAQJSK kernel $K_{\mathrm{HAQJS}}^{\mathrm{A}}$ is computed based on the density matrices of the CTQW evolving on the Hierarchical Transitive Aligned Adjacency
Matrices, that are transformed from the original graph structures (i.e., the original adjacency matrices). $K_{\mathrm{HAQJS}}^{\mathrm{A}}$ focuses more on reflecting hierarchical topological information of the original graphs. On the other hand, the proposed HAQJSK kernel $K_{\mathrm{HAQJS}}^{\mathrm{D}}$ is computed based on the Hierarchical Aligned Density Matrices, that are transformed from the original density matrices of the CTQW evolving on the original graph structures. $K_{\mathrm{HAQJS}}^{\mathrm{A}}$ focuses more on reflecting hierarchical quantum walk information of the original graphs.

% The HAQJSK kernels $K_{\mathrm{HAQJS}}^{\mathrm{A}}$ and $K_{\mathrm{HAQJS}}^{\mathrm{D}}$ are defined based on aligned structures,

% If the alignments are transitive, each of the density matrices can be aligned to a single common frame, for example that of the first graph.The aligned kernel on the original unaligned data is equivalent to the unaligned kernel on the new pre-aligned data.

\begin{table*}
\centering {
% \tiny
%\scriptsize
% \footnotesize
\caption{Properties of the Proposed HAQJSK Kernels}\label{Comparison}
\scriptsize
\vspace{-10pt}
\begin{tabular}{|c||c||c||c||c||c|}

  \hline
 ~Kernel Properties ~ & ~\textbf{HAQJSK}  ~      &~QJSK~\cite{DBLP:journals/pr/Bai0TH15,rossi2015measuring}~    & ~DBAK~\cite{DBLP:conf/icml/Bai0ZH15,DBLP:conf/ijcai/BaiZW0H15,DBLP:journals/pr/XuBJTZL21}~ &  ~R-conK~\cite{DBLP:conf/aaai/XuTJ21,DBLP:conf/aaai/SchulzWW22}, \ etc~ &  ~GGK~\cite{DBLP:journals/jmiv/BaiH13,DBLP:conf/icml/JohanssonJDB14,DBLP:journals/pr/XuJBXL18}~\\ \hline \hline

 ~Positive Definite~                     & ~$\mathrm{Yes}$~  &~ $\mathrm{No}$~   & ~$\mathrm{No}$~ &  ~$\mathrm{Yes}$~&  ~$\mathrm{Yes}$~ \\  \hline

 ~Reduce Tottering~                      & ~$\mathrm{Yes}$~  &~ $\mathrm{Yes}$~  & ~$\mathrm{-}$~  &  ~$\mathrm{-}$~  &  ~$\mathrm{-}$~ \\  \hline

 ~Structural Alignment~                  & ~$\mathrm{Yes}$~  &~ $\mathrm{Yes}$~  & ~$\mathrm{Yes}$~&  ~$\mathrm{No}$~ &  ~$\mathrm{No}$~ \\  \hline

 ~Transitive Alignment~                  & ~$\mathrm{Yes}$~  &~ $\mathrm{No}$~   & ~$\mathrm{No}$~ &  ~$\mathrm{No}$~ &  ~$\mathrm{No}$~ \\  \hline

 ~Capture Local Information~             & ~$\mathrm{Yes}$~  &~ $\mathrm{Yes}$~  & ~$\mathrm{Yes}$~&  ~$\mathrm{Yes}$~&  ~$\mathrm{No}$~ \\  \hline

 ~Capture Global Information~            & ~$\mathrm{Yes}$~  &~ $\mathrm{Yes}$~  & ~$\mathrm{No}$~ &  ~$\mathrm{No}$~ &  ~$\mathrm{Yes}$~ \\  \hline

  ~Hierarchical Alignment~   & ~$\mathrm{Yes}$~  &~ $\mathrm{No}$~   & ~$\mathrm{No}$~ &  ~$\mathrm{-}$~ &  ~$\mathrm{-}$~ \\  \hline

\end{tabular}
}
\begin{tablenotes}
\item[1] $-$: indicate that these kernels do not refer to this problem.
\end{tablenotes}
\vspace{-15pt}
\end{table*}
\subsection{Discussions of the Proposed HAQJSK Kernels}\label{s3:3}

The proposed HAQJSK kernels have a number of theoretical properties that are not available for existing state-of-the-art graph kernels, explaining the effectiveness of the proposed kernels. These properties are shown in Table~\ref{Comparison} and briefly discussed as follows.

First, unlike the existing Depth-based Alignment Kernels (DBAK)~\cite{DBLP:conf/icml/Bai0ZH15,DBLP:conf/ijcai/BaiZW0H15,DBLP:journals/pr/XuBJTZL21} that can not guarantee the transitivity between aligned vertices, the proposed HAQJSK kernels can encapsulate transitive vertex correspondence information into the kernel computation process. This is because the vertex correspondence information is identified by aligning each graph to a common set of prototype representations, i.e., only the vertices aligning to the same prototype representation will be considered as aligned vertices. Thus, the proposed HAQJSK kernels not only guarantee the positive definiteness that is not available for these DBAK kernels, but also reflect more precise kernel-based similarity measures. Moreover, unlike the DBMK kernels that only identify correspondence information between original graph structures, the proposed HAQJSK kernels can reflect hierarchical correspondence information by aligning the graphs to the hierarchical prototype representations. Hence, only the proposed HAQJSK kernels can reflect multi-scale structure information of graphs.

Second, unlike the family of R-convolution Kernels (R-conK)~\cite{haussler99convolution} that compromise to use substructures of limited sizes and only capture local structural information of graphs, the proposed HAQJSK kernels can capture global structural information by measuring the QJSD between the von Neumann entropies of global graph structures. Thus, the proposed HAQJSK kernels can reflect richer global structural information than the R-conK kernels. Furthermore, since the von Neumann entropy of a graph is computed associated with the CTQW that can reflect intrinsic structural information of graphs, the proposed can reflect more complicated topological information of graphs than the R-conK kernels.

Third, unlike the existing Quantum Jensen-Shannon Kernels (QJSK)~\cite{DBLP:journals/jmiv/BaiH13,DBLP:journals/pr/Bai0TH15,rossi2015measuring} that are also defined based on the QJSD measure associated with the CTQW, only the proposed HAQJSK kernel can encapsulate the transitive vertex alignment information to compute the required composite structure of the QJSD. Hence, only the proposed HAQJSK kernels can be considered as a kind of transitive alignment kernels, simultaneously guaranteeing the permutation invariant and the positive definiteness. Moreover, the proposed HAQJSK kernels are defined based on the hierarchial transitive aligned structures, that are transformed from the original graphs structures through the hierarchical correspondence information. By contrast, the QJSK kernels are defined based on the original graph structures. Thus, only the proposed HAQJSK kernels can reflect hierarchical structure information of graphs.

Fourth, unlike the Global Graph Kernels~\cite{DBLP:conf/icml/JohanssonJDB14,DBLP:journals/pr/XuJBXL18} that focus on capturing global structure information of graphs, the proposed HAQJSK kernels not only reflect global structure information through the CTQW, but also encapsulate local structure information through the hierarchical correspondence information of local vertices. In other words, the proposed HAQJSK kernels have better trade-off between the global and local structure information.

Finally, note that, it has be shown that some existing R-convolution graph kernels based on classical random walks~\cite{DBLP:conf/icml/KashimaTI03} and Weisfeiler-Lehman subtree patterns~\cite{shervashidze2010weisfeiler} suffer from the notorious tottering problem~\cite{DBLP:journals/pr/Bai0TH15}. This drawback is due to the fact that the substructures employed by these R-convolution kernels may contain multiple paths or edges connected by the same vertex pairs, causing redundant information and influencing the effectiveness of these R-convolution kernels. By contrast, the CTQW can theoretically reduce the tottering problem through the quantum interference between vertices. Thus, the proposed HAQJSK kernels can reduce the tottering problem through the CTQW. On the other hand, although some R-convolution kernels can overcome the tottering problem based on the backtrackless or non-backtrack paths, e.g., the non-Backtrack Path Kernel~\cite{DBLP:journals/tnn/AzizWH13}, the Shortest Path Kernel~\cite{DBLP:conf/icdm/BorgwardtK05}. Unfortunately, the backtrackless or non-backtrack paths of these R-convolution kernels can only reflect local structure information. By contrast, the proposed HAQJSK kernels not only reduce the tottering problem, but also simultaneously capture global and local structure information.

\begin{table*}
\centering {
% \tiny
% \scriptsize
 \footnotesize
\vspace{-0pt}
\caption{Information of the Graph Datasets}\label{T:GraphInformation}
\vspace{-10pt}
\begin{tabular}{|c|c|c|c|c|c|c|}
\hline
~Datasets ~        & ~MUTAG  ~  & ~PPIs~    & ~CATH2~     & ~PTC(MR)~& ~GatorBait~  & ~BAR31  ~    \\ \hline \hline

~Max \# vertices~  & ~$28$~     & ~$218$~   &  ~$568$~    & ~$109$~  & ~$545$~      & ~$220$~      \\ \hline

~Mean \# vertices~ & ~$17.93$~  & ~$109.63$~&  ~$308.03$~ & ~$25.56$~& ~$348.72$~   & ~$95.42$~    \\  \hline

~Mean \# eges~     & ~$19.79$~  & ~$531.50$~&  ~$1254.8$~ & ~$25.96$~& ~$796.11$~   & ~$94.59$~    \\  \hline

~\# graphs~        & ~$188$~    & ~$219$~   &  ~$190$~    & ~$344$~  & ~$100$~      & ~$300$~      \\ \hline

~\# vertex labels~ & ~$7$~      & ~$-$~     &  ~$-$~      & ~$19$~   & ~$-$~       & ~$-$~       \\ \hline

~\# classes~       & ~$2$~      & ~$5$~     &  ~$2$~      & ~$2$~    & ~$30$~       &  ~$20$~      \\ \hline

~Description~      & ~Bio~      & ~Bio~     &  ~Bio~      &  ~Bio~   & ~CV~         & ~CV~         \\ \hline\hline

~Datasets ~         & ~BSPHERE31~  & ~GEOD31~  & ~IMDB-B~  & ~IMDB-M~   & ~RED-B~     & ~COLLAB~      \\ \hline \hline

~Max \# vertices~   & ~$227$~      & ~$380$~   & ~$136$~   & ~$89$~     & ~$3782$~    & ~$492$~     \\ \hline

~Mean \# vertices~  & ~$99.83$~    & ~$57.24$~ & ~$19.77$~ & ~$13.00$~  & ~$429.62$~  & ~$74.49$~       \\  \hline

~Mean \# eges~      & ~$56.58$~    & ~$99.01$~     & ~$96.53$~ & ~$65.93$~  & ~$497.75$~  & ~$2457.50$~       \\  \hline

~\# graphs~         & ~$300$~      & ~$300$~   &  ~$1000$~ & ~$1500$~   & ~$2000$~    & ~$5000$     \\ \hline

~\# vertex labels~  & ~$-$~        & ~$-$~     &  ~$-$~    & ~$-$~      & ~$-$~       & ~$-$~     \\ \hline

~\# classes~        &  ~$20$~      &  ~$20$~   & ~$2$~     &  ~$3$~     & ~$2$~       & ~$2$~        \\ \hline

~Description~       & ~CV~         & ~CV~      &  ~SN~     & ~SN~       & ~SN~        & ~SN~  \\ \hline
\end{tabular}

\begin{tablenotes}
\item[1] $-$: indicate that these datasets do not have labels, and we employ the vertex degrees as the labels.
\end{tablenotes}

} \vspace{-5pt}
\end{table*}

\begin{table*}
\centering {
%\tiny
% \scriptsize
 \footnotesize
\caption{Graph Kernels for Comparisons.}\label{T:GKInfor}
\vspace{-10pt}
\begin{tabular}{|c|c|c|c|c|c|}

  \hline
 ~Kernel Methods      ~& ~Kernel Frameworks~ & ~Aligned~              & ~Transitive~            & ~Structure Patterns~       & ~Computing Models~   \\ \hline \hline

 ~\textbf{HAQJSK(A)}~ & ~Information Theory~ & ~Yes~                  & ~Yes~                   & ~Global Structures~        & ~Quantum Walks~     \\
 ~\textbf{HAQJSK(D)}~ & ~ ~                  & ~   ~                  & ~   ~                   & ~Local(Vertices)~          & ~             ~     \\ \hline

  ~QJSK~\cite{DBLP:journals/pr/BaiH14B}~              & ~Information Theory~ & ~No~                   &  ~No~                  & ~Global(Entropy)~        & ~Quantum Walks~    \\ \hline

  ~ASK~\cite{DBLP:conf/icml/Bai0ZH15}~               & ~Information Theory~ & ~Yes~                  &  ~No~                   & ~Local(Vertices)~          & ~Quantum Walks~    \\
  ~  ~                                               & ~R-convolution~      & ~   ~                  &  ~ ~                    & ~Local(Subtrees)~          & ~             ~    \\ \hline

  ~JTQK~\cite{DBLP:conf/pkdd/Bai0BH14}~             & ~Information Theory~  & ~No~                   &  ~No~                   & ~Global(Entropy)~          & ~Quantum Walks~     \\
  ~                                   ~             & ~R-convolution~       & ~  ~                   &  ~  ~                   & ~Local(Subtrees)~          & ~             ~     \\ \hline

  ~GCGK~\cite{DBLP:journals/jmlr/ShervashidzeVPMB09}~             & ~R-convolution~       & ~No~                   &  ~No~                   & ~Local(Subgraphs)~         & ~Classical~     \\ \hline
%  ~ ~                & ~R-convolution~       & ~No~                   &  ~No~                   & ~               ~          & ~             ~     \\ \hline

  ~WLSK~\cite{DBLP:journals/jmlr/ShervashidzeSLMB11}~             & ~R-convolution~       & ~No~                   &  ~No~                   & ~Local(Subtrees)~          & ~Classical~     \\   \hline

~CORE WL~\cite{DBLP:conf/ijcai/NikolentzosMLV18}~            & ~R-convolution~       & ~No~                   &  ~No~                   & ~Local(Subtrees)~          &~Classical~                       \\   \hline

  ~SPGK~\cite{DBLP:conf/icdm/BorgwardtK05}~             & ~R-convolution~       & ~No~                   &  ~No~                   & ~Local(Paths)~             & ~Classical~    \\  \hline

  ~CORE SP~\cite{DBLP:conf/ijcai/NikolentzosMLV18}~          & ~R-convolution~       & ~No~                   &  ~No~                   & ~Local(Paths)~             & ~Classical~                       \\  \hline

  ~PMGK~\cite{DBLP:conf/aaai/NikolentzosMV17}~             & ~R-convolution~       & ~Yes~                  &  ~No~                   & ~Local(Vertices)~          & ~Classical~                      \\ \hline

  ~SPEGK~\cite{DBLP:journals/pr/XuBJTZL21}~           & ~Information Theory~ & ~Yes~                  &  ~No~                   & ~Local(Vertices)~          & ~Classical~               \\ \hline

\end{tabular}
} \vspace{-5pt}
\end{table*}

\subsection{The Computational Complexity}

Assume the set of graphs $\mathbf{G}$ has $\mathcal{N}$ graphs, each graph in $\mathbf{G}$ has $n$ vertices and $m$ edges, computing the proposed HAQJSK kernels over $\mathbf{G}$ needs three computational steps, i.e., a) compute the DB representations rooted at each vertex, b) construct the hierarchical prototype representations, c) compute the hierarchical correspondence matrix, and d) compute the QJSD between hierarchical transitive aligned structures. The first computational step relies on evaluating the shortest path between vertices of each graph, and thus needs time complexity $O(\mathcal{N}n\log n +\mathcal{N}mn)$. The second computational step relies on the $\kappa$-means clustering method and requires time complexity $O(HIMn)$, where $H$ is the greatest hierarchical level of the parameter $h$ defined previously, $M$ is number of the $1$-level prototype representations (i.e., $M=|\mathbf{{P}}^{1,k}|$ defined previously), and $I$ is the iteration number for $\kappa$-means. The third computational step relies on computing the $h$-level hierarchical correspondence matrix between each graph to the $h$-level hierarchical prototype representations, and thus needs time complexity $O(HM\mathcal{N}n)$. The fourth computational step depends on on the spectral decomposition of the CTQW, and thus needs time complexity $O(\mathcal{N}^2n^3)$. Thus, the whole time complexity is $O( \mathcal{N}n\log n +\mathcal{N}mn  + HIMn   + H\mathcal{N}Mn  +\mathcal{N}^2 n^3)$. Because $N\gg M$, $N\gg I$, $N\gg H$, and $n^2\gg m$, the resulting time complexity of the HAQJSK kernels are $$O(\mathcal{N}^2n^3),$$ indicating that the proposed kernels have a polynomial time.

% requires time complexity $O(N^2n^3)$. Specifically, computing the proposed kernels

% \subsection{Experiments on Graph Classification}

\begin{table*}
\centering {
%\tiny
% \scriptsize
 \footnotesize
\caption{Classification Comparisons using Graph Kernels.}\label{T:ClassificationGK}
\vspace{-10pt}
\begin{tabular}{|c|c|c|c|c|c|c|}

  \hline
 ~Datasets~& ~MUTAG  ~         & ~PPIs~           & ~CATH2~            & ~PTC(MR)~       & ~GatorBait~     & ~BAR31~           \\ \hline \hline

 ~\textbf{HAQJSK(A)}~  & ~$85.83\pm0.72$~& ~$\textbf{89.71}\pm0.54$~&  ~$83.47\pm0.88$~ & ~$\textbf{62.35}\pm0.51$~ & ~$\textbf{20.00}\pm0.84$ &~$68.00\pm0.60$~   \\

 ~\textbf{HAQJSK(D)}~ & ~$86.33\pm0.81$~& ~$86.28\pm0.41$~ & ~$\textbf{87.89}\pm0.35$~  & ~$59.05\pm0.62$~ & ~$\textbf{22.80}\pm0.89$ &~$71.70\pm0.61$~   \\ \hline

  ~QJSK~     & ~$82.72\pm0.44$~  & ~$65.61\pm0.77$~ &  ~$71.11\pm0.88$~ & ~$56.70\pm0.49$~ & ~$9.00\pm0.89$ &~$30.80\pm0.61$~   \\ \hline

  ~ASK~      & ~$87.50\pm0.65$~  & ~$80.14\pm0.73$~ &  ~$78.52\pm0.67$~ & ~$56.22$~        & ~$7.50\pm0.74$ &~$\textbf{73.10}\pm0.67$~   \\ \hline

  ~JTQK~   & ~$85.50\pm0.55$~  & ~$88.47\pm0.47$~ &  ~$68.70\pm0.69$~ & ~$58.50\pm0.39$~ & ~$11.40\pm0.52$ &~$60.56\pm0.35$~   \\ \hline

  ~GCGK~   & ~$81.66\pm2.11$~  & ~$46.61\pm0.47$~ &  ~$73.68\pm1.09$~ & ~$52.26\pm1.41$~ & ~$8.40\pm.83$   &~$22.96\pm0.65$   \\ \hline

  ~WLSK~   & ~$82.88\pm0.57$~  & ~$88.09\pm0.41$~ &  ~$67.36\pm0.63$~ & ~$58.26\pm0.47$~ & ~$10.10\pm0.61$ &~$58.53\pm0.53$~   \\   \hline

~CORE WL~  & ~$87.47\pm1.08$~  & ~$-$~            & ~$-$~             & ~$59.43\pm1.20$~ &~$-$             & ~$-$~             \\   \hline

  ~SPGK~   & ~$83.38\pm0.81$~  & ~$59.04\pm0.44$~ &  ~$81.89\pm0.63$~ & ~$55.52\pm0.46$~ & ~$9.00\pm0.75$  &~$55.73\pm0.44$~   \\  \hline

  ~CORE SP~& ~$\textbf{88.29}\pm1.55$~  & ~$-$~            &  ~$-$~            & ~$59.06\pm0.93$~ & ~$-$            &~$-$~              \\  \hline

  ~PMGK~   & ~$86.67\pm0.60$~  & ~$-$~            &  ~$-$~            & ~$60.22\pm0.86$~ & ~$-$            &~$-$~              \\ \hline

  ~SPEGK~  & ~$86.35$~          & ~$84.13$~        &  ~$83.58$~        & ~$56.79$~        & ~$14.40$       &~$70.08$~         \\ \hline \hline

 ~Datasets~& ~BSPHERE31~       & ~BEOD31~         & ~IMDB-B~         & ~IMDB-M~          & ~RED-B~         & ~COLLAB~           \\ \hline \hline

~\textbf{HAQJSK(A)}~&~$58.40\pm0.66$~   &~$45.26\pm0.74$~  & ~$\textbf{73.50}\pm0.45$~ & ~$50.08\pm0.20$   & ~$\textbf{90.93}\pm0.12$~  & ~$\textbf{79.20}\pm0.17$\\

~\textbf{HAQJSK(D)}~&~$\textbf{61.60}\pm0.53$~   &~$\textbf{47.53}\pm0.34$~  & ~$72.57\pm0.31$~ & ~$49.30\pm0.39$~  & ~$\textbf{89.50}\pm0.17$~  & ~$\textbf{78.82}\pm0.14$ \\ \hline

~QJSK~     & ~$24.80\pm0.61$~  & ~$23.73\pm0.66$~ &  ~$62.10$~        & ~$43.24$~        & ~$-$              &~$-$~   \\ \hline

~ASK~      & ~$60.30\pm0.44$~  & ~$46.21\pm0.69$~ &  ~$63.57$~        & ~$42.81$~        & ~$-$              &~$-$~   \\ \hline

  ~JTQK~   &~$46.93\pm0.61$~   &~$40.10\pm0.46$~  &~$72.45\pm0.81$~  & ~$50.33\pm0.49$~  & ~$77.60\pm0.35$~& ~$76.85\pm0.40$\\ \hline

  ~GCGK~   &~$17.10\pm0.60$~   &~$15.30\pm0.68$~  &~$65.87\pm0.98$~  & ~$45.42\pm0.87$~  & ~$77.34\pm0.18$~& ~$-$\\ \hline

  ~WLSK~   &~$42.10\pm0.68$~   &~$38.20\pm0.68$~  &~$71.88\pm0.77$~  & ~$49.50\pm0.49$~  & ~$76.56\pm0.30$~& ~$77.39\pm0.35$\\   \hline

~CORE WL~  & ~$-$~             & ~$-$~            &~$74.02\pm0.42$~  & ~$\textbf{51.35}\pm0.48$~  & ~$78.02\pm0.23$ & ~$-$~  \\   \hline

  ~SPGK~   &~$48.20\pm0.76$~   & ~$38.40\pm0.65$  &~$71.26\pm1.04$~  & ~$51.33\pm0.57$~  & ~$84.20\pm0.70$~& ~$58.80\pm0.20$\\  \hline

  ~CORE SP~&~$-$~              &~$-$~             &~$72.62\pm0.59$~  & ~$49.43\pm0.42$~  & ~$90.84\pm0.14$~& ~$-$           \\  \hline

  ~PMGK~   & ~$-$~             &~$-$~             & ~$68.53\pm0.61$~ & ~$45.75\pm0.66$~  & ~$82.70\pm0.68$~& ~$-$  \\ \hline

  ~SPEGK~  & ~$57.36$~         & ~$43.57$~        &  ~$-$~            & ~$-$~            & ~$-$              &~$-$~         \\ \hline
\end{tabular}

\begin{tablenotes}
\item[1] $-$: indicate that some methods were not evaluated on corresponding datasets by the original or other authors by now.
\end{tablenotes}

} \vspace{-5pt}
\end{table*}

\begin{table*}
\vspace{-0pt}
\centering {
%\tiny
% \scriptsize
 \footnotesize
\caption{Classification Comparisons using Deep Learning Methods.}\label{T:ClassificationGCNN}
\vspace{-10pt}
\begin{tabular}{|c|c|c|c|c|c|c|}

  \hline
 ~Datasets~  & ~MUTAG  ~        & ~PTC(MR)~         & ~IMDB-B~         & ~IMDB-M~        & ~RED-B~           & ~COLLAB~  \\ \hline \hline
 ~\textbf{HAQJSK(A)}~ & ~$85.83\pm0.72$~ & ~$\textbf{62.35}\pm0.51$~  & ~$\textbf{73.50}\pm0.45$~ & ~$50.08\pm0.20$ & ~$\textbf{90.93}\pm0.12$~  & ~$\textbf{79.20}\pm0.17$\\

 ~\textbf{HAQJSK(D)}~ & ~$86.33\pm0.81$~ & ~$59.05\pm0.62$~  & ~$72.51\pm0.31$~ & ~$49.30\pm0.39$~& ~$\textbf{89.50}\pm0.17$~  & ~$\textbf{78.82}\pm0.14$ \\ \hline

  ~DGCNN~  & ~$85.83\pm1.66$~  & ~$58.59\pm2.47$~ & ~$70.03\pm0.86$~& ~$47.83\pm0.85$~ & ~$76.02\pm1.73$~  & ~$73.76\pm0.49$\\ \hline

  ~PSGCNN~ & ~$\textbf{88.95}\pm4.37$~  & ~$62.29$~       & ~$71.00\pm2.29$~& ~$45.23\pm2.84$~ & ~$86.30\pm1.58$~  & ~$72.60\pm2.15$\\ \hline

  ~DCNN~   & ~$66.98$~        & ~$58.09\pm0.53$~ & ~$49.06\pm1.37$~& ~$33.49\pm1.42$~ & ~$-$~             & ~$52.11\pm0.71$\\ \hline

  ~DGK~    & ~$82.66\pm1.45$~ & ~$57.32\pm1.13$~ & ~$66.96\pm0.56$~& ~$44.55\pm0.52$~ & ~$78.30\pm0.30$~  & ~$73.09\pm0.25$\\ \hline

  ~AWE~    & ~$87.87\pm9.76$~ & ~$-$~           & ~$73.13\pm3.28$~& ~$\textbf{51.58}\pm4.66$~ & ~$82.97\pm2.86$~  & ~$70.99\pm1.49$\\ \hline

%  ~HO-GCN~ & ~$86.10$~          & ~$60.90$~       & ~$\textbf{74.20}$~       & ~$49.50$~        & ~$-$~             & ~$-$\\ \hline

\end{tabular}

\begin{tablenotes}
\item[1] $-$: indicate that some methods were not evaluated on corresponding datasets by the original or other authors by now.
\end{tablenotes}

} \vspace{-18pt}
\end{table*}

\section{Experiments}\label{s4}

In this section, we compare the classification performance of the proposed kernels to both state-of-the-art graph kernels and deep learning methods on standard benchmark graph datasets.

\subsection{Benchmark Dtasets}
We evaluate the proposed kernels on twelve standard graph datasets abstracted from bioinformatics (Bio), computer vision (CV) and social networks (SN), respectively. Specifically, the Bio and SN datasets can be found in~\cite{KKMMN2016}. The CV datasets are introduced in the references~\cite{DBLP:conf/dgci/BiasottiMMPSF03,DBLP:conf/cvpr/EscolanoHL11}. Details of the statistical information for these datasets are exhibited in Table.\ref{T:GraphInformation}.

\subsection{Comparisons with Graph Kernels}
\textbf{Experimental Setups:} We compare the classification performance of the proposed HAQJSK kernels with some classical graph kernels. These kernels are 1) the Quantum Jensen-Shannon Kernel (QJSK) associated with the CTQW~\cite{DBLP:journals/pr/BaiH14B}, 2) the Aligned Subtree Kernel (ASK)~\cite{DBLP:conf/icml/Bai0ZH15} with the highest subtree layer $50$, 3) the Jensen-Tsallis q-difference Kernel (JTQK)~\cite{DBLP:conf/pkdd/Bai0BH14} with $q=2$ and the subtrees of height $10$, 4) the Graphlet Count Graph Kernel (GCGK)~\cite{DBLP:journals/jmlr/ShervashidzeVPMB09} with graphlet of size $4$, 5) the Weisfeiler-Lehman subtree kernel (WLSK)~\cite{DBLP:journals/jmlr/ShervashidzeSLMB11} with the subtrees of height $10$, 6) the WLSK kernel associated with Core Variants (CORE WL)~\cite{DBLP:conf/ijcai/NikolentzosMLV18}, 7) the Shortest Path Graph Kernel (SPGK)~\cite{DBLP:conf/icdm/BorgwardtK05}, 8) the SPGK kernel associated with Core Variants (CORE SP)~\cite{DBLP:conf/ijcai/NikolentzosMLV18}, 9) the Pyramid Match Graph Kernel (PMGK)~\cite{DBLP:conf/aaai/NikolentzosMV17}, and 10) the depth-based Second-order R{\'{e}}nyi Entropy Graph Kernel (SREGK)~\cite{DBLP:journals/pr/XuBJTZL21}. More detailed properties of the kernels for comparisons are shown in Table~\ref{T:GKInfor}.

For the proposed HAQJSK kernels, we set the hierarchical parameter $H$ as $5$ and thus compute $5$ kernel matrices for each datasets. Moreover, we set the number of the $1$-level prototype representations $\mathbf{{P}}^{h,k}$ ($h=1$) as $256$ (i.e., $|\mathbf{{P}}^{1,k}|=256$), because the number $256$ is larger than the average graph sizes of most graph datasets. Specifically, we perform the $10$-fold cross-validation strategy to compute the classification accuracy through the C-Support Vector Machine (C-SVM)~\cite{ChangLinSVM2001} associated with the graph kernels. For each kernel, we employ the optimal C-SVMs parameters and repeat the experiment for 10 times on each dataset. We show the average classification accuracy ($\pm$ standard error) in Table~\ref{T:ClassificationGK}. For some alternative kernels, the experimental results are directly from the original references or the comprehensive review paper~\cite{DBLP:journals/entropy/ZhangWW18}, following the same experimental setups with our HAQJSK kernels.

\textbf{Experimental Results and Analysis:} In terms of the classification performance, it is clear that the proposed HAQJSK kernels can significantly outperform the alternative graph kernels on nine of the twelve datasets. Although, the classification accuracies of the proposed HAQJSK kernels are lower than those of some alternative kernels on the MUTAG, BAR31, and IMDB-M datasets, the HAQJSK kernels are still competitive to these kernels and better than other alternative kernels.

The reasons of the effectiveness for the proposed HAQJSK kernels are due to the advantages that have been well explained in Section~\ref{s3:3}. First, the proposed HAQJSK kernels can either capture the complicated characteristics of global graph structures from the CTQW or reflect local structural information through the correspondence information between vertices. By contrast, the alternative graph kernels ASK, JTQK, GCGK, WLSK, CORE WL, SPGK, CORE SP and PMGK are the instances of R-convolution graph kernels, and are defined based on substructures of small sizes. As a result, these alternative kernels can only reflect restricted local structural information. Second, as a quantum R-convolution kernel, although the JTQK kernel can also reflect comprehensive global structural information through the CTQW. Similar to the other alternative R-convolution kernels, the JTQK kernel mainly focuses on measuring the isomorphism between substructures and ignores the structural correspondence information between the substructures. Thus, all the alternative R-convolution kernels, including the JTQK kernel, can not reflect precise kernel-based similarity measure between graphs. Third, although the QJSK kernel is also defined based on the QJSD measure between global graphs associated with the CTQW. The QJSK kernel cannot integrate the transitive correspondence information between graphs. As a result, the QJSK kernel is not a positive definite kernel, and can not reflect precise similarity measure between graphs. Moreover, the QJSK kernel focuses on the global structure information through the CTQW evolve on the graphs, lacking local structural information. Fourth, although the ASK and the SPEGK kernels can also identify the correspondence information between graphs. These two kernels can not guarantee the transitivity between the aligned vertices. Finally, note that, the proposed HAQJSK kernels are not attributed kernels and cannot accommodate the vertex attributed information. By contrast, the ASK, JTQK, WLSK and CORE WL kernels are attributed kernels. But the proposed HAQJSK kernels can still outperform these attributed kernels. In summary, the proposed HAQJSK kernels are more effective than the alternative kernels, and the strategy of the hierarchical structures associated with the CTQW can really improve the performance of graph kernels.

\subsection{Comparisons with Graph Deep Learning}

\noindent\textbf{Experimental Setups:} We compare the classification performance of the proposed HAQJSK kernels with some graph deep learning methods. These methods are 1) the Deep Graph Kernel (DGK)~\cite{DBLP:conf/kdd/YanardagV15}, 2) the Diffusion Convolutional Neural Network (DCNN)~\cite{DBLP:conf/nips/AtwoodT16}, 3) the PATCHY-SAN based Convolutional Neural Network for graphs (PSGCNN)~\cite{DBLP:conf/icml/NiepertAK16}, 4) the Deep Graph Convolutional Neural Network (DGCNN)~\cite{DBLP:conf/aaai/ZhangCNC18}, and 5) the Anonymous Walk Embeddings based on feature driven (AWE)~\cite{DBLP:conf/icml/IvanovB18}. Since the alternative deep learning follow the same experimental setup with our proposed kernels, we directly repot the classification accuracies from the original corresponding references in Table~\ref{T:ClassificationGCNN}. Note that the PSGCNN model can also adopt the edge attributes. However, since most datasets and other methods do not leverage, we only report the results of the PSGCNN model with vertex features.

\textbf{Experimental Results and Analysis:} In terms of the classification performance, it is clear that the proposed HAQJSK kernels can significantly outperform the alternative graph deep learning methods on most of the datasets. In fact, the C-SVMs associated with the graph kernel can be seen as a kind of methods based on the shallow learning strategy, that may have lower performance than the deep learning methods. But the proposed HAQJSK kernels still have better classification performance than these deep learning methods. This may due to the fact that some of the deep learning methods (i.e., the graph convolutional network models DGCNN, PSGCNN and DCN) are theoretically related to the Weisfeiler-Lehman (WL) isomorphism test of the WLSK kernel~\cite{DBLP:conf/aaai/ZhangCNC18}, since they all rely on the information propagation between adjacent vertices. As a result, similar to the WLSK kernel, these graph convolution network models may also suffer from the tottering problem. By contrast, the CTQW of the proposed HAQJSK kernels can reduce the tottering problem. Thus, this evaluation again indicates that the strategy of the hierarchical structures associated with the CTQW can improve the performance of graph kernels.

% \subsection{Parameter Evaluation of the Proposed Kernels}

\section{Conclusions and Future Works}\label{s5}
In this work, we propose a family of HAQJSK kernels based on measuring the QJSD between hierarchical transitive aligned structures of graphs associated with the CTQW. The proposed HAQJSK kernels not only overcome the shortcoming of ignoring structural correspondence information arising in classical R-convolution kernels, but also simultaneously reflect richer local and global structural information than the R-convolution kernels that only reflect local structural information on substructures. Moreover, unlike most existing alignment kernels, the proposed HAQJSK kernels can guarantee the transitivity between the correspondence information. Thus, the proposed HAQJSK kernels not only positive definite kernels, but also reflect more precise similarity measure between graphs. The experimental evaluations demonstrate the effectiveness. Our future work is to develop the proposed HAQJSK kernels on step further, and integrate the vertex label information into the kernel computation, resulting  new attributed HAQJSK kernels.

\section*{Acknowledgments}
This work is supported by the National Natural Science Foundation of China under Grants T2122020, 61976235, and 61602535. Corresponding Author: Lixin Cui.

%  (cuilixin@cufe.edu.cn).

% was supervised by Dr. Lu Bai and Dr. Lixin Cui for his M.sc degree and

%-------------------------------------------------------------------------

\balance

%-------------------------------------------------------------------------
%\nocite{ex1,ex2}
%\bibliographystyle{latex12}
%\bibliography{XBib}

\bibliographystyle{IEEEtran}
\bibliography{example_paper}

\end{document}